\documentclass[smallcondensed,a4paper]{llncs} 

\usepackage{graphicx}
\usepackage{multirow} 
\usepackage{arydshln} 
\usepackage{subfigure} 

\usepackage{url} 
\usepackage{xcolor} 

\begin{document}

\title{Creation and Evaluation of Datasets for Distributional Semantics Tasks in the Digital Humanities Domain}

\titlerunning{Distributional Semantics Datasets in Digital Humanities}

\author{Gerhard Wohlgenannt\inst{1} \and Ariadna Barinova\inst{1} \and Dmitry Ilvovsky\inst{2} \and Ekaterina Chernyak\inst{2}}

\institute{Faculty of Software Engineering and Computer Systems, ITMO University, St. Petersburg, Russia \\
\and National Research University, Higher School of Economics, Moscow, Russia}



\maketitle

\begin{abstract}

Word embeddings are already well studied in the general domain, usually trained on large text corpora, and have been evaluated for example 
on word similarity and analogy tasks, but also as an input to downstream NLP processes. 
In contrast, in this work we explore the suitability of word embedding technologies in the specialized digital humanities domain. 
After training embedding models of various types on two popular fantasy novel book series, 
we evaluate their performance on two task types: term analogies, and word intrusion. To this end, we manually construct test datasets with domain experts. 
Among the contributions are the evaluation of various word embedding techniques on the different task types, 
with the findings that even embeddings trained on small corpora perform well for example on the word intrusion task.
Furthermore, we provide extensive and high-quality datasets in digital humanities for further investigation, 
as well as the implementation to easily reproduce or extend the experiments.


\end{abstract}

\section{Introduction}
\label{sec:intro}

Distributional semantics (DS) is based on the notion that the meaning of words is given by their linguistic context~\cite{Harris1954,Firth1957}.
Early work includes for example Grefenstette~\cite{Grefenstette1994},
whereas more recent surveys of using vector space models of semantics are provided by~\cite{Turney2010,Erk2012,Clark2012}.
Within DS, word embeddings have shown state-of-the-art performance on word similarity tasks, but also for more sophisticated operations
like word analogies and word intrusion. 
The performance of word embeddings using huge text corpora text has been demonstrated in the previous work~\cite{mikolov2013w2v}.

Language technologies (LT) are increasingly adapted in the field of digital humanities, 
which is also reflected by a rising number of scientific events\footnote{For example: \url{https://www.clarin-d.net/en/current-issues/lt4dh}}.
In the last couple of years, word embeddings have become a very popular and successful tool for language modeling. 
Word embeddings transform the vocabulary of a text corpus into a continuous and low-dimensional vector representation. 
Word embeddings have two major applications: 
they are used to solve different tasks in semantics, and serve as input component to various natural language processing (NLP) tasks~\cite{mikolov2013w2v,ghanny2016}.
In this research we study the suitability of different types of word embeddings as a LT tool on various task types
in the digital humanities (DH) domain, and with comparably small corpus sizes.

With regards to the task types, in DS a few standard tasks are often applied to evaluate word vector representations and word embeddings:
i) word similarity, ii) word analogy, iii) word dissimilarity (in the form of \emph{word intrusion}). 
Our main goal is to adopt these tasks to the DH domain, to create respective datasets (provide ground truth), and 
evaluate the datasets with LT tools to establish baselines for future research.

The research questions addressed in this work can be grouped into two classes:
\begin{description}
    \item[Performance of Word Embeddings for Digital Humanities:] Here, the general focus is the suitability of word embedding-based LT tools
    in the DH domain. Some of the main questions are: Given the specific domain, and the small corpus size -- how good is the performance 
    on the basic tasks of (i) analogy and (ii) word intrusion? Which embedding techniques work well on which task type? Which settings, for example 
    regarding word window size, are best suited for which task type?
    How do word embeddings compare against state-of-the-art baseline methods? 

    \item[Small corpora and Named Entities (NE):] Furthermore, many interesting points arise regarding the specifics of the datasets and domain, 
    for example how entities compare against other nouns in embedding models, or what impact term frequency of the terms in the tasks has on 
    the evaluation metrics.

\end{description}

To address the research questions and the two task types, we focus on the domain of literary texts, and analyze certain aspects of two well-known fantasy novel book series, 
namely ``A Song of Ice and Fire'' by George R.~R.~Martin, and ``Harry Potter'' by Joanne K.~Rowling. For both novels we train word embedding
models with LT tools such as word2vec~\cite{mikolov2013w2v}, GloVe~\cite{pennington2014glove}, fastText~\cite{bojanowski2016} or LexVec~\cite{SalleIV16a}.

For the two tasks, firstly, we manually create high-quality datasets for two fantasy novel book series.
In our experiments we also distinguish between uni-gram and n-gram (phrases) models, esp. n-grams help to cover various named entities
in the novels. 
Many proper names, for example \emph{Many Faced God} can only be tracked without ambiguity using n-grams.
As we have two task types here, namely \emph{analogy} and \emph{word intrusion}, two book series, and models for uni-grams and
n-grams, we finally compile eight distinct datasets for evaluation. The total number of evaluation questions is $31474$. 
The analogy tasks include test data for example on the relations \emph{husband::wife}, \emph{sigil-animal::house}, and many others.
Also the datasets for the word intrusion task are grouped into thematic sections.
We use a number of baselines, PPMI, stock embedding models, etc. 

The results of the evaluations of the various word embedding techniques on the datasets are found in Section~\ref{sec:eval}.
As many of the tasks are very hard for word embedding models trained on small corpora, esp. the analogy tasks,
we do not always expect high numbers in accuracy, but aim to provide easily reproducible baselines for future work on the given test datasets.

In order for other researchers to compare their methods on the test data, we make all datasets, word embedding models, and
evaluation code available online\footnote{\url{https://github.com/gwohlgen/digitalhumanities_dataset_and_eval}}.
Furthermore, by making the dataset creation and evaluation processes simple and transparent, we aim to provide the
basis for an extension of the datasets, the addition of new datasets, and the advancement of the evaluation code base.

The remainder of the paper is structured as follows: 
After Section~\ref{sec:related} introduces related work, 
we discuss various aspects of the task types, the word embedding and baseline methods, the dataset creation and the implementation of the system
in Section~\ref{sec:methods}. Section~\ref{sec:eval} presents the evaluation setup and evaluation results 
for the two book series and the two task types, and finally Section~\ref{sec:concl} concludes the paper with a summary and contributions.

\section{Related Work}
\label{sec:related}

We relate our work to two trends in Natural Language Processing (NLP): 
Firstly, fantasy books recently became a popular subject of research in the Digital Humanities field.
Secondly, word embeddings, being the main DS tool, are also widely studied and used in various fields.
From DS we adopt two tasks: word analogy, word intrusion. 

Several factors contribute to the recent popularity of fantasy novels as source for analysis in NLP:
i) such books often have a linear timeline suitable for timeline and storyline extraction~\cite{laparra2015timelines},
ii) they feature a profound amount of direct speech for dialogue~\cite{flekova2015personality} and social network analysis~\cite{bonato2016mining}. 

Modern Distributional Semantics is built under the assumption that the sense of the word can be represented as a dense vector (otherwise called embedding) 
and the similarity between two words can be computed as the cosine between two corresponding vectors~\cite{nayak2016evaluating,baroni2014don}. 
There are numerous techniques to generate such vectors. 
Count-based methods date back to 
Latent Semantic Analysis (LSA)~\cite{deerwester1990indexing} and Singular Value Decomposition on positive PMI word-context matrices 
as well as other weighting schemes and dimensionality reduction techniques~\cite{baroni2014don}. 
Predictive models became very popular in recent years for language modeling and feature learning, esp.~since the work of Mikolov et al.~\cite{mikolov2013w2v} on the word2vec toolkit in 2013. 
Other well-known word embedding types including GloVe~\cite{pennington2014glove}, fastText~\cite{bojanowski2016} or LexVec~\cite{SalleIV16a} have received a lot of attention even outside of the NLP community. 
The variety of embedding models have been tested and evaluated on a set of common tasks and datasets. 
According to Nayak et al.~\cite{nayak2016evaluating,baroni2014don} these tasks usually include synonym detection, word analogies, and word dissimilarity. \\

Standard datasets for word similarity are for example MEN~\cite{Bruni2014}, SimLex-999~\cite{Hill2015} and WordSim353~\cite{finkelstein2001placing}.
The analogy task was introduced in Mikolov et al.~\cite{mikolov2013w2v}. Finley et al.~\cite{finley2017analogies} lists several datasets for the analogy task and their comparison. 
The word intrusion task was introduced originally to measure the cohesiveness of a latent topic in Chang et al.~\cite{chang2009reading}, 
than used in Gensim~\cite{gensim} to solve ``odd one out'' exercises for language learners.

A lot of work has been done on comparing different word embeddings models. 
Ghannay et al.~\cite{ghanny2016} did extensive evaluations to compare different kinds of embeddings, such as word2vec CBOW and skip-gram, GloVe, CSLM and word2vec-f (see Section~\ref{sec:methods} for details on the algorithms). 
The various methods and tools perform very differently depending on the task.
GloVe had the best performance in analogical reasoning, and CBOW/skip-gram were best at word similarity tasks.
Lison and Kutuzov~\cite{lison2017redefining} study parameter settings for context windows in training word2vec skip-gram models evaluated 
both on word similarity and analogy tasks. In their experiments larger word windows have a positive effect in analogy tasks, whereas 
the opposite effect is observed for word similarity. Furthermore, word windows across sentences sometimes were beneficial.

Faruqui et al.~\cite{faruqui2016problems} 
identify problems associated with evaluating embedding models only on word similarity tasks and suggested to rather conduct task- and domain-specific evaluations. \\ 

Linzen~\cite{Linzen2016} discusses potential pitfalls of the vector offset method of analogy, and presents baselines to improve the utility of vector space evaluations. 
We apply the suggested baselines in our evaluations.

As our work focuses on small sized corpora, the work of Sahlgren and Lenci~\cite{Sahlgren2016emnlp}, who study the effect of corpus size on model performance, is very relevant.
They evaluate various DS model types (co-occurrence, PPMI, SVD, word2vec, random indexing) mainly on word similarity tasks. 
PPMI and TSVD outperform word2vec models for a corpus size comparable to ours (1M token) with a very small word window (2).
Herbelot and Baroni~\cite{Herbelot2017emnlp} look at tiny data size, they guess embeddings for new words from only a couple of sentences using a very high learning rate.
In this situation it is very important that the sample sentences are significant of word usage. Based on these ideas, we include experiments
which update pre-trained corpora with the book corpora with different learning rates.

The majority of terms in our analogy and word intrusion datasets are named entities denoted with proper names. 
In contrast to other types of noun phrases, proper names are not well studied in DS~\cite{Herbelot2015darcy,Gupta2017starsem}.
In recent years, some works appeared that investigate the specifics of entities. Gupta et al.~\cite{Gupta2015emnlp} predict discrete referential attributes of 
entities from DS models. Boleda et al.~\cite{Boleda2017eacl} study entities and concepts in DS, 
esp.~the respective instance-of and hypernymy relations. While our datasets are not directed at specific types of relations, 
they contain instance-of analogy tasks.
Herbelot~\cite{Herbelot2015darcy} contextualizes concepts with local entity information, and point out important characteristics of proper names 
(uniqueness, instantiation and individuality).
Gupta et al.~\cite{Gupta2017starsem} predict semantic relations between entities collected from Freebase with the analogy method and a feed-forward neural network, 
and analyze factors for task difficulty, such as 1:n relations and relations with many instances.

In our work we draw from different aspects of the discussed related work.
We create domain-specific datasets in the fantasy novel domain for two popular book series, 
and those datasets target the discussed typical evaluations tasks: word analogy and word intrusion (a task strongly related to word similarity).

In Wohlgenannt et al.~\cite{wohlgenannt2016lt4dh} we presented and evaluated methods to extract social networks using
co-occurrence statistics and word embedding models from a novel book series. The proposed work builds on
parts of Wohlgenannt et al.~\cite{wohlgenannt2016lt4dh}, esp.~the application of word embeddings in the digital humanities domain,
but extends previous research with specific relation types like analogical reasoning and word intrusion,
the manual creation and provision of datasets, and is applied to multiple fantasy novels.

\section{Methods and Implementation}
\label{sec:methods}

This section presents the details on the methods used to evaluate the application of word embeddings
in the domain of literary fiction. First, we introduce the two types of tasks in a more formal and detailed way. 
Then follows a description of the techniques applied, 
starting with the baseline methods such as PPMI, and afterwards we specify the various types of word embeddings.
Next, we discuss the methods applied for dataset creation using domain experts.
And finally, the section gives an overview of the accompanying implementation of the work discussed in this publication.

\subsection{Task Types}

In the course of this work we evaluate two task types on the two book series. 
The tasks cover \emph{analogies} and \emph{word intrusion} between arbitrary terms and concepts in the books.

\subsubsection{Analogies}
\label{sec:m:ana}
The two task types are frequently used for the evaluation of word embedding techniques, esp.~the task type of analogies. 
The initial implementation of word2vec~\cite{mikolov2013w2v} includes a test dataset with various test groups and a script to apply
a given model to the test dataset. 

The classical example for analogies is: \texttt{king} -
\texttt{man} + \texttt{woman} = \texttt{?}, where the correct solution is \texttt{queen}.
Using embedding models, this task can be solved with vector arithmetics. 
As stated, the word2vec toolkit contains a general domain test dataset with tasks such as \\
\emph{capital\_city$_a$, country$_a$ :: capital\_city$_b$, ?} \\
or \emph{adjective$_a$, superlative$_a$ :: adjective$_b$, ?}. \\
In this task, a result candidate is selected from the vocabulary of the model with vector arithmetics.
As any term from the complete vocabulary is a possible candidate, this task
is quite hard to solve. The candidate is selected simply as the word vector closest
to the vector which results from combining the three input vectors (with subtraction and addition).
Similarity in vector space is typically calculated with the cosine similarity. 

For this task, and the next, we manually created high-quality test datasets for the two book series
(see Section~\ref{sec:m:dsc} for details on dataset creation).

\subsubsection{Word Intrusion}
\label{sec:m:dm}

    Secondly, the \emph{word intrusion} task simply selects an outlier from a list of input terms.
    More formally, we created test datasets, where each task unit consists of $4$ terms,
    where one term semantically doesn't match with the other terms. A simple example
    in the ``A Song of Ice and Fire'' world would be \texttt{Lannister Stark Theon Martell},
    where all terms, except the outlier \texttt{Theon} are names of families. 
    For task evaluation we rely on the popular Gensim library\footnote{\url{https://radimrehurek.com/gensim}}.

    This task is obviously easier than analogical reasoning, as the system only has a small selection
    of terms (in our experiments $4$ input terms) to choose from, 
    and not the whole vocabulary, like in the previous task. The random baseline is $0.25$.

    After the first set of experiments we decided to formalize the task further and make it more meaningful
    and challenging. For this reason, the dataset provides 20 wrong candidates to be mixed 
    into every triple of semantically related terms, therefore generating 20 task units per triple.
    The 20 outliers are distributed equally into four categories of task difficulty. 
    The first 5 outliers are hard to distinguish, in the second group the candidates are rather loosely semantically related to the task term,
    in third group the relation is weak, for example only being of the same term type (for example: named entity).
    The final group of 5 candidates is completely unrelated to the three task terms.
    In the evaluations we measure the total accuracy and the accuracy per difficulty category to better understand the performance 
    of the models. 
   
    To illustrate difficulty levels with an example from the dataset, we start from the related triple \texttt{Ned Robb Arya} (who are all
    members of the ``Stark'' family. In difficulty category 1 we might mix in \texttt{Theon}, who is no Stark family member,
    but lived with the family for some time. 
    In category 2, one of the outliers is \texttt{Bolton}, which is a family name and not a person, but as related to the Stark family. 
    In category 3 an example is \texttt{Harrenhal}, which only has the same term type (named entity), but no other relation. 
    And finally in category 4 an example outlier is \texttt{sword}, which is just a random term from the books.

\subsection{Baseline Methods}

To address the tasks defined in the datasets (see below), multiple techniques and combinations of methods
are feasible. In the research, we focus on the application and evaluation of different word embedding models.
The baseline methods are used in the evaluation section (Section~\ref{sec:eval}) 
to compare the results from word embeddings to these state-of-the-art techniques.

\subsubsection{PPMI}
\label{sec:ppmi}
PPMI (positive pointwise mutual information) is often put in the category of count-based methods for distributional semantics, 
in contrast to prediction-based methods like word2vec~\cite{levy2015}.
PPMI is a weighting method applied to a high-dimensional sparse matrix of vocabulary words as rows, 
and potential contexts as columns. The cells represent the association of 
a word and a context. For details on PMI and PPMI computation see~\cite{church1990} and~\cite{levy2015}.
PPMI outperforms PMI on semantic similarity tasks and is regarded a state-of-the-art method
for distributional similarity~\cite{levy2015}.
In our implementation we used the PyDSM library\footnote{\url{https://github.com/jimmycallin/pydsm}}
for PPMI.

\subsubsection{Stock Word Embeddings}
\label{sec:stock-fasttext}
The book series we investigated (``A Song of Ice and Fire'' and ``Harry Potter'') are part of mass culture, which is reflected by their presence in Wikipedia. 
We used the English language model trained on Wikipedia available online\footnote{\url{https://github.com/facebookresearch/fastText/blob/master/pretrained-vectors.md}} as a baseline
for models trained on the book corpora.

\subsubsection{Analogy Baseline: ONLY-B}
\label{sec:only-b}
Linzen~\cite{Linzen2016} shows that the conventional vector offset method for analogy can conflate offset consistency with largely irrelevant neighborhood
structure. In an analogy relation of \emph{A:A* :: B:B*}, the simple ONLY-B baseline completely ignores \emph{A} and \emph{A*} and just 
returns the terms with highest similarity to \emph{B}. This baseline helps to distinguish success from the vector offset method from 
simple neighborhood structure. Linzen also studies other baselines, for example \emph{IGNORE-A}, which give the word most similar to $(A* + B)$.
He recommends to report ONLY-B and IGNORE-A for analogy tasks.

\subsection{Word Embedding Methods}

As already stated, the evaluation of methods based on word embeddings in the context
of literary fiction is one of the main goals of this publication.
In general, word embedding models have been shown to be very successful on many NLP and language modeling tasks~\cite{ghanny2016}, 
and furthermore they are easy to train, apply and compare.
This subsection briefly introduces the applied word embedding methods from
a theoretical point of view.

\subsubsection{Word2vec}
\label{sec:w2v}

The word2vec~\cite{mikolov2013w2v} toolkit \footnote{\url{https://code.google.com/archive/p/word2vec}} applies two-layer neural networks.
which are trained in an unsupervised way. The input is a (typically large) corpus, 
the results are the word embeddings. Depending on the preprocessing of the corpus, users can train uni-gram
or n-gram models. Proximity in vector space corresponds to similar contexts in which words appear.
Within this low-dimensional representation, word vector length is typically in the range of of 50 to 300. 
The number of dimensions is a training parameter, and depends on the task at hand.
There are two model architectures to create the continuous vector representations: 
continuous bag-of-words (CBOW) or continuous skip-gram. 
With CBOW, the model predicts the current word by using a window of surrounding words. 
Using skip-gram, the model predicts the surrounding window of context words by using the current word.
As also confirmed in the upcoming evaluations, the tuning of word2vec hyperparameters has a large
impact on performance~\cite{Sahlgren2016emnlp}. 
Other hyperparameters include the maximum size of the word windows, and the minimum frequency of a word to be included in the word matrix.
With Gensim, vector space models can be updated. We experiment with that feature 
when updating a model trained on Wikipedia (\emph{text8}) corpus with the book corpora.

\subsubsection{GloVe}
\label{sec:glove}

Another well-known contender in the word embeddings field is GloVe~\cite{pennington2014glove}, 
which also learns continuous vector representations of words. 
In contrast to word2vec it is not a predictive, but rather a count-based model. 
GloVe applies dimensionality-reduction on a word-word co-occurrence matrix.
The training objective is to learn word vectors such that their dot product 
equals the logarithm of the words' probability of co-occurrence.
In this research we used the GloVe implementation from Stanford university\footnote{\url{http://nlp.stanford.edu/projects/glove, GloVe version 1.2}}
for model training.


\subsubsection{fastText}
\label{sec:m:fasttext}
FastText~\cite{bojanowski2016} is based on the skip-gram model, but also makes use of word morphology information in the training process,
representing each word as a bag of character n-grams.
By using sub-word information, and constructing word vectors as the sum of character n-gram vectors, 
fastText\footnote{\url{https://fasttext.cc}} can supply better vectors for rare words, and even out-of-vocabulary words.
The authors report state-of-the-art performance on word similarity and analogy tasks comparing against deep learning methods.

%

%
%
%

\subsubsection{LexVec}
\label{sec:m:lexvec}

LexVec~\cite{SalleIV16a, Salle2018} is a word embedding method which uses low-rank, weighted factorization of the PPMI matrix and combines characteristics of methods such as word2vec and GloVe.
It tackles a well-known shortcoming of PPMI by assigning heavier penalties for errors on frequent co-occurrences.
LexVec\footnote{\url{https://github.com/alexandres/lexvec}} was shown to perform well on word similarity and semantic analogy tasks, but struggles on syntactic analogies.

\subsection{Dataset Creation}
\label{sec:m:dsc}
\label{m:adm}

In Wohlgenannt et al.~\cite{wohlgenannt2016lt4dh} we used crowdsourcing 
for the \emph{Social Network Extraction} task to create a gold standard.
The analogies and word intrusion tasks are harder to describe to crowd workers and require manual inspection, 
therefore we decided to rely on domain experts to create the datasets.

We created eight gold datasets manually, resulting from any combination of these three ingredients:
\begin{description}
    \item[Task Type:] Analogies and word intrusion.
    \item[Book Series:] ``A Song of Ice and Fire'' (ASOIF), and ``Harry Potter'' (HP).
    \item[Term Type:] Uni-gram versus n-gram. 
\end{description}

In the process of dataset creation, the starting points were the book series wikis\footnote{\url{http://awoiaf.westeros.org/index.php?title=Special:Categories}} 
\footnote{\url{http://harrypotter.wikia.com/wiki/Main_Page}}.
The categories, lists and other items in the wikis helped to collect data and ideas on task units both for word intrusion and for analogy relation tasks, but finally it was 
not possible to extract gold data from the wikis automatically, but it was necessary to manually filter, edit and extend the data to ensure high dataset quality.
\footnote{Manual refinement became necessary for a number of reasons:
i) The frequency of entities mentioned in wikis is often below our minimum frequency (5 occurrences).
ii) In case of the uni-gram dataset construction, ambiguity between entities is a problem, which necessitates manual refinement.
iii) Finally, as the word2vec toolkit generates a low number of n-grams, the dataset creators had to use exactly those n-grams,
in order for the models to find the terms in the vocabulary.
}

\subsection{Implementation}
\label{sec:impl}

In this section we focus specifically on the basic implementation of the task types 
\emph{analogies} and \emph{word intrusion}, in order to allow for others
to reproduce and extend the work with little effort.  

We only give a brief overview here, as all code and documentation, 
the word embedding models, and the datasets are available online\footnote{\url{https://github.com/gwohlgen/digitalhumanities_dataset_and_eval}}.
The results can be reproduced by cloning the repository and running the evaluation
scripts. 

The currently available implementation has two main components, 
the creation of datasets and the use of those datasets to evaluate the models. 
Regarding the creation of datasets, starting from a simple format to describe task sections
and task units (see Github), the \texttt{create\_questions.py} script creates the evaluation dataset as all permutations 
of the input definitions.

For system evaluation, there are two main scripts for the \emph{analogies} and \emph{word intrusion}
tasks (\texttt{analogies\_evaluation.py} and \texttt{doesnt\_match\_evaluation.py}).
They iterate over the word embedding models defined in the configuration file,
run the task units from the datasets, and collect and aggregate all evaluation results.

The implementation makes use of the Gensim library~\cite{gensim} for loading the models, and performing
the two basic task types. 
For the \emph{word intrusion} task, the analysis of the results based on task type and task difficulty 
is conducted with Python pandas\footnote{\url{http://pandas.pydata.org/}}. 


\section{Evaluation}
\label{sec:eval}

In the evaluation section we address the research questions posed in the introduction section with the tools and methods
introduced in Section~\ref{sec:methods}.
First, we discuss on evaluation setup, esp.~the book corpora and word embedding models, 
then give a quick overview of the datasets, 
and in Section~\ref{sec:eval_results} we present the results of experiments for the analogies and word intrusion tasks.

\subsection{Evaluation Setup}
\label{sec:gsetup}

The main ingredient to all downstream tasks is the underlying text corpus. 
Here we present some basic facts about the two fantasy novel books series,
and about the settings used to train models on those corpora.

\subsubsection{Text Corpora} 

\paragraph{A Song of Ice and Fire}

The first set of experiments was conducted with the plain text version of the first four books of the ``A Song of Ice and Fire'' (ASOIF) book series by George R. R. Martin. 
ASOIF is a fantasy novel book series. The action takes place in a fictional medieval-like universe, which also includes elements of magic. 
Although the number of characters is immense, 
there are around to 30-40 main characters which communicate and interact throughout the first four books of the series.  
Narration is mostly linear with occasional flashbacks, and in first person perspective. 
However each chapter has a distinct character in focus, the story is told from different viewpoints. This way,
the story unfolds in different parts of the ASOIF world in parallel.  
The raw books amount to 7.0MB of plain text, 1.3M tokens, and contain 204 chapters with a mostly chronological storyline. 
There are 121098 sentences in total.

\paragraph{Harry Potter}
To contrast and verify some of the results of the experiments on ASOIF, we chose to repeat the evaluations on 
``Harry Potter'' (HP) by Joanne K.~Rowling (all books).
The story takes place at Hogwarts School of Witchcraft and Wizardry and describes the adventures of the main character Harry Potter and his friends. 
The complexity of the world in HP and the number of characters that intensively interact with each other throughout the story is generally lower 
than in ASOIF.
The HP series is of similar size as the ASOIF text, with 6.5MB file size and 1.1M tokens.

We chose those two books series, as they are a) very popular and well known, which might support the reuse of the datasets,
b) the small corpus sizes and specific use of language allow to address the research question whether such corpora
can be sufficient to utilize word embedding models.

\paragraph{Corpus Preprocessing}

Corpus preprocessing is kept to a minimum and includes removal of punctuation and quotation symbols, and sentence splitting. 
The word2vec built-in tools for preprocessing are used for creating the n-gram corpora.

\subsubsection{Grid Search for Model Parameter Settings}

As mentioned, results from DS models often depend more on hyperparameter settings than model type~\cite{Sahlgren2016emnlp}.
Here, we provide a quick survey of the result of a grid search for hyperparameter optimization of word2vec and FastText
models for both \emph{word intrusion} and \emph{analogies}, following the approach of Antoniak and Mimno~\cite{antoniak2018evaluating}.
The goal was gain some understanding on the influence of model parameters on the individual tasks,
and to find suitable parameter settings for the upcoming evaluations.

\begin{figure}[h!]
\centering
\begin{subfigure}
\centering
\includegraphics[width=0.95\linewidth]{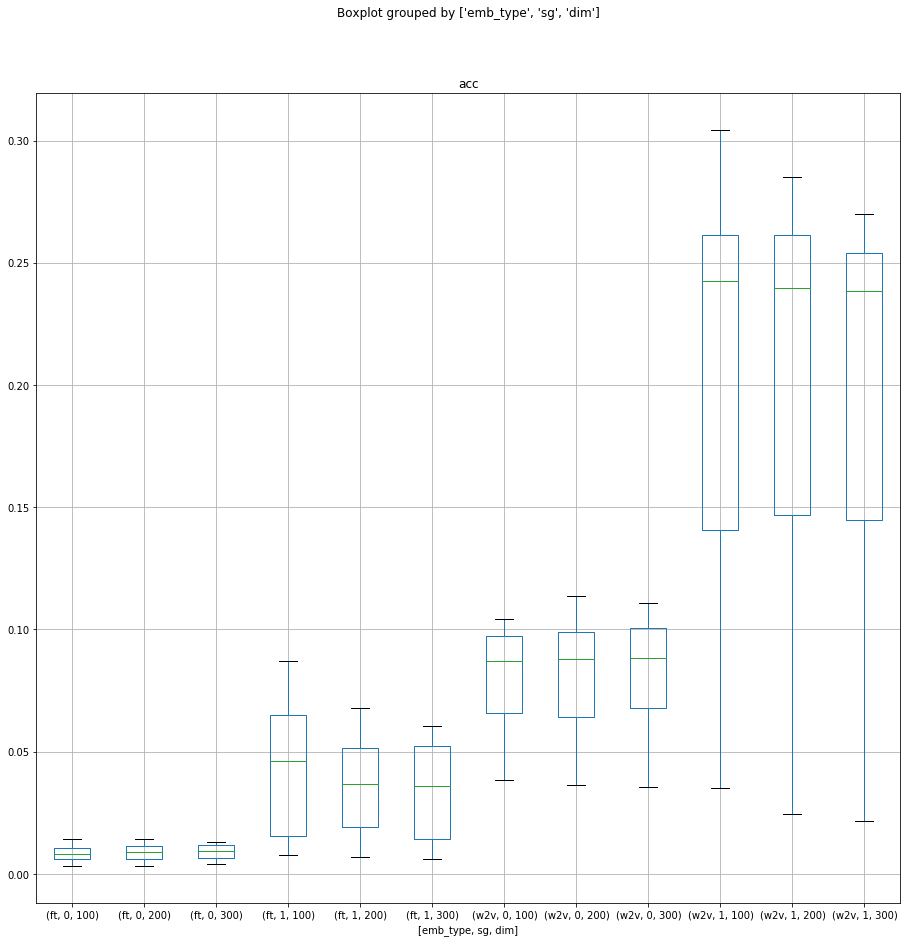}
\end{subfigure}
\caption{The boxplot shows the spread of accuracy values for analogy task for ASOIF, with different model types and vector sizes.}
\label{figure:boxplots}
\end{figure}

The search was implemented with Gensim, we also use their naming conventions for \emph{parameters}:
\begin{itemize}
	\item Embedding model: word2vec (w2v) or FastText (ft)
	\item Vector dimensionality: 100, 200 or 300
	\item Training algorithm: Skip-gram (sg=1) or CBOW (sg=0);
	\item Window size: 1, 2, 3, 5, 7, 9, 11, 13, or 15
	\item Number of negative contexts for negative sampling: 5, 10, or 15
\end{itemize}

Combining these parameters, we trained 648 models for both ASOIF and HP, which were evaluated for the analogies and word intrusion tasks.

Figure~\ref{figure:boxplots} presents the spread of accuracy values in the analogy task for ASOIF averaged over window sizes and number of negative contexts. 
It shows the results of FastText CBOW, FastText SG, word2vec CBOW and word2vec SG regarding vector size (100, 200, 300).
Overall, word2vec models clearly outperform FastText on the analogy task, and vector dimensionality has low effects on accuracy. Similarity, 
the parameter \emph{negative sampling} had little effect on average.

\begin{figure}[h!]
\begin{subfigure}
\centering
\includegraphics[width=0.5\linewidth]{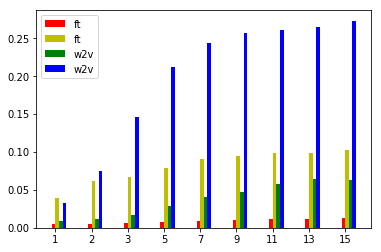}
\end{subfigure}
\begin{subfigure}
\centering
\includegraphics[width=0.5\linewidth]{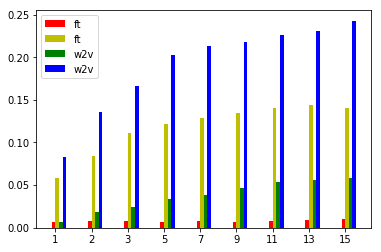}
\end{subfigure}
\caption{Effects of window size on accuracy values for analogy tasks for ASOIF (left) and HP (right).}
\label{figure:window}
\end{figure}

Figure~\ref{figure:window} shows the impact of \emph{word window sizes} on analogy task accuracy. Results indicate, that small windows (esp.~if smaller than $5$ words) 
lead to low performance in our task setup. This confirms previous work which recommends small windows for syntactic tasks, but larger windows for semantic tasks~\cite{Linzen2016}.



To save space we do not include the results for the word intrusion task, but those are in line with the results presented, although gaps between FastText and word2vec are smaller.
The overall result is that word2vec is favorable to FastText, and that the Skip-Gram algorithm performs better than CBOW.



\subsubsection{Models Trained}
\label{sec:e:mt}

Based on the grid search on model hyperparameters, we decided to use following configurations : a) As model dimensionality has little impact on results,
we train all models with 300-dimensions (a standard setting), in order to highlight and distinguish the impact of other parameters.
b) Skip-gram is the algorithm of choice for word2vec and FastText according to the grid search, but still we train also some models with CBOW to gain
insight into potential differences in performance. c) Larger word windows seem to provide better results, therefore we decided to apply a window 
of 12 words in most models, instead of the (Gensim) default of 5 words.

In order to compare and evaluate the performance of various word embedding types in the upcoming evaluations,
above considerations lead to the training of the following models on the two book corpora (all models available on github):

\begin{description}
        \item[w2v-default:] This is a word2vec model trained with the Gensim default settings, which are CBOW, word window of 5 words,
        negative sampling (5 samples), 5 epochs. The only change made to the defaults: 300-dim vectors instead of 100-dim.

        \item[w2v-ww12-i15-ns:] word2vec with skip-gram algorithm, window size of 12, 15 epochs, and negative sampling with 15 \emph{noise words}, 300 dimensions.
        \item[w2v-ww12-i15-hs:] Like the previous model, but with hierarchical softmax instead of negative sampling. 
        \item[w2v-CBOW:] Same settings like \emph{w2v-ww12-i15-ns}, but using CBOW instead of skip-gram method.

        \item[GloVe:] Using the defaults. Only changes: 300-dim vectors instead of 50-dim, window size 12.

        \item[fastText-default:] We used the default settings, which are CBOW algorithm, 5 epochs, word window size of 5, negative sampling (5 samples). Exception from default:
        300-dim. vectors.

        \item[fastText-ww12-i15-ns:] Same settings like in \emph{w2v-ww12-i15-ns}.

        \item[LexVec:] We used the default settings, except: 25 epochs, window size 5 (instead of default: 2).
\end{description}

On github, depending on the book series used, the models are prefixed with \texttt{asoif\_} or \texttt{hp\_}.
The models above were trained on uni-gram corpora, and can be used for the uni-gram datasets.
For the n-gram datasets and evaluation tasks, we provide n-gram models on github, too. 

\begin{figure*}[htb]
\centering
{\centering \resizebox*{0.70\textwidth}{!}{
\includegraphics{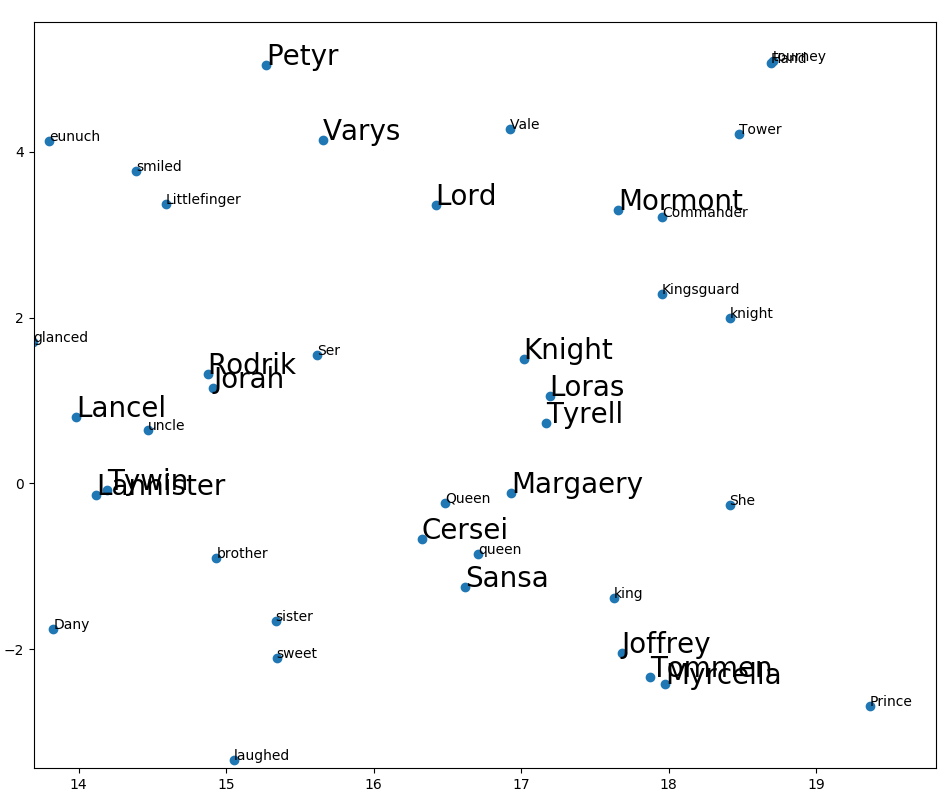}
}} \caption{\label{fig:screen_tsne} 
A small part of the ASOIF GloVe model vector space (reduced to two dimensions with t-SNE).}
\end{figure*}

Figure~\ref{fig:screen_tsne} shows a visualization of a small fragment of the ASOIF GloVe model and illustrates
the basic concept of word embeddings. The two-dimensional visualization was created with the help of 
t-SNE~(\cite{TSNE}) for dimensionality reduction.

\subsection{Datasets}
\label{sec:eval_datasets}

%

In Section~\ref{m:adm} we introduced the dataset creation process. The combination of two book series, 
two task types, and two term/model versions (uni-gram and n-gram) results in eight datasets.
Those eight datasets were created manually with the goal of high quality.
Table~\ref{tab:ds} gives an overview of the dataset statistics. The datasets are found in the github 
repository\footnote{\url{https://github.com/gwohlgen/digitalhumanities_dataset_and_eval}} in the \texttt{datasets} 
folder\footnote{All files start with \texttt{questions\_}. The rest of the file name reflects the books series (\texttt{soiaf} or \texttt{hp}),
the task type (\texttt{analogies} or \texttt{doesnt\_match}), and finally n-gram datasets are marked with \texttt{\_n-gram}.
This naming convention should make the contents of the datasets clear.}.
The individual datasets are furthermore grouped into sections of tasks of a specific type (for example finding \emph{child-to-father} relations).
In total our datasets contain 16220 questions for the ASOIF book series, and 15142 for HP, which adds
up to 31474 evaluation units. In general it was easier to create uni-gram datasets, as word2vec phrase detection creates only 
a comparably low number of n-grams.

With regards to the distinction between kinds and proper nouns~\cite{Herbelot2015darcy}, the majority of terms in the tasks
are named entities. For example, in the ASIOF word intrusion dataset only 7\% of terms are \emph{kinds} in the uni-gram dataset,
and only 9\% of n-gram terms are not entities. For HP, the respective numbers are 17\% and 13\%.

\begin{table}
\caption{Overview of datasets used for the Analogies and Word Intrusion tasks. Cells contain the number of questions (task units) and number of task categories in parentheses.}
\label{tab:ds}
\begin{center}
\begin{tabular}[h]{cc|cc} 
\hline\noalign{\smallskip}

 \textbf{Book Series} & \textbf{Dataset-Type} & \textbf{Uni-gram} & \textbf{N-Gram}   \\  

\noalign{\smallskip}\hline\noalign{\smallskip}

 \multirow{2}{*}{ASOIF}    &  Analogies      &  2848  (8)   &  192 (2)  \\ 
                           &  Word Intrusion &  11180 (13)  & 2000 (7)  \\ \hline 
 \multirow{2}{*}{HP}       &  Analogies      &   4822 (20)  &   92 (7)  \\ 
                           &  Word Intrusion &   8420 (20)  & 1920 (7)  \\ 
\noalign{\smallskip}\hline
\end{tabular}
\end{center}
\end{table}

\subsection{Evaluation Results}
\label{sec:eval_results}

\subsubsection{Analogies Task}

    In the analogy task, the input is a triple of terms, which can be read as $x_1$ is to $x_2$, what $y_1$ is to $y_?$.
    For example, \emph{man} is to \emph{king} what \emph{woman} is to $X$. Or, in the ASOIF universe, regarding relations of sigil-animals to houses, 
    \emph{kraken} is to \emph{Greyjoy} what \emph{lion} is to $X'$ (correct: \emph{Lannister}).
    The system has to guess the correct answer from the whole vocabulary. The vocabulary size in the models used here is between $11K$ to $60K$ terms.
    Given the comparably small corpus size, this task is very hard. 
    But on a deeper level, this task is hard as characters (or even abstract concepts) have a multitude of relations
    between each other.
    This characteristic also made dataset creation challenging for the analogy task. Additionally,
    esp.~for the uni-gram datasets, there is the problem of ambiguities, as there are many re-occurring names and nicknames of entities in the large ASOIF universe.
    Furthermore, relations change over time. 

    Every task is defined by four terms, the three input terms, and the correct answer to the task.
    The tasks are split into various sections, for example predicting \emph{child-to-father} relations, \emph{houses-to-their-seats}, etc.
    For every task unit (question), Gensim uses vector arithmetic to calculate the candidate term, and compares it to the correct solution given in the dataset. 

    \begin{table}[htb]
    \caption{ASOIF Analogies dataset (uni-grams): Accuracy of various word embedding models on selected analogy task groups, and total accuracy.}
    \label{tab:e:a1}
    \begin{center}
    \begin{tabular}{l|l|l|l|l|l|l}
    \hline\noalign{\smallskip}

          \multirow{2}{*}{Task Section}  & first-    & child- & husband- & name-    & houses- & \multirow{2}{*}{Total} \\
                                         & last-name & father & wife     & location & seats &                          \\ \hline 

    \noalign{\smallskip}\hline\noalign{\smallskip}
    Number of tasks:       & 2368 & 180 & 30 & 168 & 30 & 2848  \\ \hline

    \noalign{\smallskip}\hline\noalign{\smallskip}
 w2v-default                &  7.98  & 8.33 & 3.33 & 1.19  & 56.67  &  7.87   \\ 

 w2v-ww12-i15-hs            &  37.8  & 4.44 & 3.33 & 16.07 & 30.0   & \textbf{33.57}   \\ 

 w2v-ww12-i15-ns            &  30.74 & 0.0  & 0.0  & 4.17  & 13.33  & 26.12   \\ 

 w2v-CBOW                   &  9.54  & 7.22 & 6.67 & 8.33  & 46.67  & 9.59    \\ 

 GloVe                      &  36.23 & 3.89 & 0.0  & 0.0   & 30.0   & 30.86   \\ 
 fastText-default           &  0.93  & 6.67 & 6.67 & 1.19  & 6.67   & 1.4     \\ 
 fastText-ww12-i15-ns       &  32.73 & 1.67 & 0.0  & 4.76  & 16.67  & 27.88   \\ 

 LexVec                     &  0.04  & 8.89 & 0.0  & 0.0   & 0.0    & 0.6     \\  
 Text8 w2v updated          &  36.05 & 2.63 & 6.67 & 5.41 & 43.33 & 32.29  \\ \hdashline
 Baseline: ONLY-B           &  31.12 & 0.0 & 0.0 & 0.0 & 0.0 & 25.81  \\ 

    \noalign{\smallskip}\hline

    \end{tabular}
    \end{center}
    \end{table}

    In this section, we first discuss the results for the ASOIF dataset, and later compare those to the HP book series.
    Table~\ref{tab:e:a1} presents the evaluation results for the ASOIF analogy dataset.
    Due to space limitations, we selected the results for five dataset sections, and
    the aggregated results. 
    The total number of task units is $2848$, the by far largest section are \emph{firstname-lastname} analogy questions.
    The best results were provided by a word2vec model (\emph{w2v-ww12-i15-hs}), but on this task, also \emph{GloVe} is competitive.
    Both the \emph{w2v-default} setting with its CBOW model and a small word window of 5, and the word2vec \emph{CBOW}
    model with a larger window seem unsuited for this task configuration. 
    FastText models yield lower accuracy than word2vec overall.
    Generally the accuracy is low with around 33.6\%;
    the reasons for the difficulty of the task were already discussed above. 
    The ONLY-B baseline~\cite{Linzen2016} used with the \emph{w2v-ww12-i15-hs model} provides very interesting results, it gives 0\% accuracy for most categories,
    but for the \emph{firstname:lastname} task, the results are high for such a simple baseline. 
    This shows, that the characters last names tend to be very close to their first names in vector space.

    With respect to Gupta et al.~\cite{Gupta2017starsem}, who identify 1:n relations, and relations with many instances,
    as harder in analogy experiments, we analyzed our analogy task categories. Some tasks like \emph{houses:seats}
    contain only 1:1 relations, but we did not find clear evidence for differences in performance in the uni-gram datasets.
    In the n-gram datasets the accuracy on \emph{name:nickname} (1:1) is high -- however, a deeper analysis in future work
    is necessary to draw conclusions.


    \begin{table}[htb]
    \caption{Harry potter Analogies dataset: Accuracy of various word embedding models on selected analogy task groups, and total accuracy.}
    \label{tab:e:a2}
    \begin{center}
    \begin{tabular}{l|l|l|l|l|l}
        \hline\noalign{\smallskip}

\multirow{2}{*}{Task Section} & first-    & child-   & husband-    & wizard-    & \multirow{2}{*}{Total} \\ 
                              & last-name & father   & wife        & faculty    &                        \\ 

    \noalign{\smallskip}\hline\noalign{\smallskip}
    Number of tasks:       & 2390 & 224 & 72 & 566 & 4822  \\ \hline

    \noalign{\smallskip}\hline\noalign{\smallskip}

w2v-default                &  7.66  & 8.48  & 13.89 & 7.77  & 13.69  \\ 
w2v-ww12-i15-hs            &  41.26 & 12.05 & 44.44 & 34.28 & \textbf{32.43} \\ 
w2v-ww12-i15-ns            &  33.43 & 5.36  & 34.72 & 22.44 & 22.4   \\ 
w2v-CBOW                   &  12.05 & 9.38  & 11.11 & 20.85  & 15.47 \\ 

 GloVe                     &  38.74 & 5.36  & 18.06 & 17.67 & 27.71  \\ 
 fastText-default          &  1.55  & 0.45  & 1.39  & 0.35  & 1.33  \\ 
 fastText-ww12-i15-ns      &  31.88 & 7.14  & 34.72 & 27.74 & 22.6  \\ 

 LexVec                    &  1.3   & 0.45  & 1.39  & 0.18  & 0.84   \\ 
 Text8 w2v updated         &  22.5  & 1.56  & 16.67 & 23.9  & 20.77  \\ \hdashline
 Baseline:  ONLY-B         & 49.08  & 6.25  & 44.44 & 41.52 & 30.42  \\

    \noalign{\smallskip}\hline

    \end{tabular}
    \end{center}
    \end{table}

    The independent evaluation of analogical reasoning with a dataset for the Harry Potter books supports most observations made on the ASOIF dataset.
    Table~\ref{tab:e:a2} presents the results for the $4790$ task units.
    Again, the setting \emph{w2v-ww12-i15-hs} with hierarchical softmax performs best, followed by \emph{GloVe}. 
    Our results confirm the findings of Ghannay et al.~\cite{ghanny2016}, which report good performance of GloVe on analogical reasoning tasks.
    The score of CBOW-based models remains low.

    The evaluation results show large disparities between different task groups and methods.
    It is notable, that even within the same task group there can be large differences between the 
    two book series. For example, the section \emph{husband-wife} has very poor results 
    in the ASOIF books, while for the HP books results are much better.

    The ONLY-B baseline provides good results for a number of categories in the HP dataset, which shows that
    for many of these relation pairs the terms are neighbors in vector space. Finally, we didn't measure benefits from 
    initializing the model with pretraining on a \emph{text8} Wikipedia dataset (.

    The same set of experiments was conducted using the n-gram datasets and models trained on the phrase-annotated corpora. 
    The test datasets are much smaller, with 192 task units and 2 groups in the ASOIF dataset, and 92 task units (7 groups) for \emph{Harry Potter}.
    As the results are generally in line with the uni-gram evaluations, we won't include the full results in table form. 
    For the ASOIF n-gram dataset, tuned word2vec (\emph{w2v-ww12-i15-ns-ngram}) performed best with around 16.2\%. 
    On HP data, \emph{w2v-default} was in front.
    Overall, average accuracy is lower for n-gram than for uni-gram datasets.

    On the basis of our findings in grid search, and in Lison and Kutuzov~\cite{lison2017redefining}, where large word windows provide
    high accuracy in the analogy task, we tested also some very large word windows sizes, as well as word windows across sentence boundaries. 
    We trained additional models with word window sizes 15, 30 and 45 for both variants, ie.~where the window can cross the sentence boundary,
    or not (default). First of all, if the context window needs to stay within the sentence, large windows have little impact, as 
    in the books few sentences contain 30 or more words. However, large windows which cross sentence boundaries improved the
    accuracy by a few percent -- depending on the model settings. In conclusion, if the embedding model is primarily used for 
    analogical reasoning, then these results should be considered during parameter search. For word similarity tasks like word intrusion
    smaller windows are preferable (see below).


\subsubsection{Word Intrusion Task}

    For the word intrusion task, the input dataset contains four terms, where three are semantically connected, and one is mixed in.
    To distinguish the correct answer, it is explicitly provided in the dataset after a separation symbol.
    The evaluation script calls Gensim to provide the outlier candidate, which is then compared to the correct answer.
    Internally, Gensim computes the mean vector from all four term vectors, and then calculates the cosine distance for each term. 
    The vector with the largest distance to the mean is selected as not matching the rest.
    The random baseline for this task is $\frac{1}{4}$ (0.25). 

    As for the analogy task, we present the evaluation results for both book series, and in combination with uni-gram and n-gram models.
    Additionally, the \emph{word intrusion} datasets include information about task difficulty, and an analysis regarding the effects 
    of term frequency on accuracy, both are empirically investigated. 

    \begin{table}[ht]
    \caption{ASOIF Word Intrusion dataset: Accuracy of various word embedding models on selected word intrusion task sections, and total accuracy.}
    \label{tab:e:dm1}
    \begin{center}
    \begin{tabular}{l|l|l|l|l|l}
    \hline\noalign{\smallskip}
          Task Section & family-siblings  & names-of-houses  & Stark clan & free cities & Total \\
            \noalign{\smallskip}\hline\noalign{\smallskip}
Number of tasks:       & 160 & 7280 & 1120 & 700 & 11180  \\ 

            \noalign{\smallskip}\hline\noalign{\smallskip}

 w2v-default                &  83.12 & 82.28 & 88.48 & 82.43 & \textbf{83.01}  \\ 
w2v-ww12-i15-hs             &  84.38 & 63.32 & 90.27 & 94.14 & 72.89  \\ 
w2v-ww12-i15-ns             &  82.5  & 67.39 & 91.16 & 90.71 & 75.93  \\ 
w2v-CBOW                    &  81.25 & 76.1  & 87.32 & 85.71 & 79.73  \\ 

 GloVe                      &  80.62 & 73.19 & 90.62 & 88.14 & 76.25  \\ 
fastText-default            &  90.62 & 73.71 & 87.41 & 63.57 & 76.06  \\ 
fastText-ww12-i15-ns        &  83.12 & 69.35 & 86.25 & 90.71 & 76.72  \\ 

 LexVec                     &  82.5  & 62.36 & 90.36 & 82.43 & 70.93  \\ 

 Text8 w2v updated          &  83.75 & 57.73 & 89.29 & 93.57 & 67.93  \\ \hdashline
 
 Baseline: PPMI             &  82.8  & 67.95 & 75.0  & 99.47 & 70.37  \\ 
 Baseline: WP FastText      &  54.37 & 51.73 & 46.07 & 36.71 & 49.69  \\ 

    \noalign{\smallskip}\hline
    \end{tabular}
    \end{center}
    \end{table}

    We start with the evaluation results of the ASOIF book series and the uni-gram dataset, which is presented in Table~\ref{tab:e:dm1}.
    The dataset includes 11180 task units in 13 sections.
    Due to the lower task complexity accuracy numbers are in the range of 70\% to 83\%, with best results for \emph{w2v-default}
    with $83.01\%$. 
    Interestingly, both for the ASOIF and for Harry Potter datasets (see Table~\ref{tab:e:dm1}) \emph{CBOW} (\emph{w2v-default}, \emph{ft-default}, and \emph{w2v-CBOW}) 
    perform surprisingly well on the word intrusion tasks, esp.~with a smaller window size, which suggests that for this task setup a smaller and more focused word context has benefits.

    In the \emph{word intrusion} task evaluations we apply \emph{PPMI} (see Section~\ref{sec:ppmi}) as state-of-the-art baseline method.
    \emph{PPMI} performance is quite good, similar to some of the word embedding models. 

    \begin{table}[ht]
    \caption{Harry Potter Word Intrusion dataset (uni-grams): Accuracy of various word embedding models on selected word intrusion task sections, and total accuracy.}
    \label{tab:e:dm1:hp}
    \begin{center}
    \begin{tabular}{l|l|l|l|l|l}
    \hline\noalign{\smallskip}
\multirow{2}{*}{Task Section} & family- & Gryffindor- & magic     & \multirow{2}{*}{professors} & \multirow{2}{*}{Total} \\ 
                              & members & members     & creatures &                             &                        \\ 
    \noalign{\smallskip}\hline\noalign{\smallskip}
Number of tasks:       & 440 & 2800 & 700 & 400 & 8420 \\ \hline

    \noalign{\smallskip}\hline\noalign{\smallskip}

 w2v-default                &  78.86 & 75.61 & 57.71 & 91.0  & 71.82  \\ 
 w2v-ww12-i15-hs            &  80.45 & 76.5 & 60.71 & 80.75  & 70.27  \\ 
 w2v-ww12-i15-ns            &  82.73 & 81.79 & 41.29 & 68.25 & 73.06  \\ 
 w2v-CBOW                   &  81.36 & 80.54 & 66.57 & 90.0  & \textbf{75.33} \\ 

 GloVe                      &  85.68 & 66.25 & 33.86 & 79.0  & 64.95  \\ 
 fastText-default           &  83.64 & 75.57 & 45.71 & 83.5  & 65.33  \\ 
 fastText-ww12-i15-ns       &  80.23 & 81.04 & 43.86 & 61.0  & 71.94  \\ 
 LexVec                     &  59.55 & 54.36 & 60.0  & 84.0  & 55.11  \\ 


Text8 w2v updated          &  86.14 & 76.14 & 26.29 & 82.75 & 61.83  \\ \hdashline
Baseline: PPMI             &  70.91 & 65.46 & 22.0  & 59.0  & 56.43  \\ 
Baseline: WP FastText &  86.82 & 45.75 & 33.43 & 78.5 & 45.4  \\

\noalign{\smallskip}\hline
    \end{tabular}
    \end{center}
    \end{table}

    Again, we contrast the ASOIF dataset evaluation with the HP dataset. The Harry Potter data includes
    8420 questions in 20 groups. Word2vec-based methods excel with about $70-75\%$ accuracy, followed by \emph{fastText}.
    The differences between task sections are wide, for the \emph{professors} group most models provide 
    high accuracy, while eg.~for \emph{magic creatures} most methods struggle.

    These differences in performance can be related to theoretical distinctions between entities and kinds
    made for example in Herbelot~\cite{Herbelot2015darcy}. Whereas entities typically have discrete attributes, e.g.
    \emph{Jaime} has blond hair, kinds (\emph{knight}) reflect a distribution of attribute values. 
    Kinds are to a large extent made up by the ``supremum'' of their instances.
    In the evaluations we saw, that intruders were easier to detect into a group of individuals than
    kinds. For example, accuracy is high for the categories \emph{professors}, \emph{archmaesters}, \emph{dragons}, \emph{persons of a family}.
    For kinds on the other hand, for example (types of) \emph{magic creatures}, types of \emph{wizard equipment}, etc., performance is lower.
    Experiments on a larger scale are necessary in future work to confirm these findings.

    As discussed in the presentation of results of the analogies tasks, we experimented also with large word windows of 15, 30, and 45 words.
    In contrast to analogical reasoning, for word intrusion large windows, which cross sentence boundaries, have a substantial and consistent negative 
    effect on accuracy. This is in line with the results on word similarity experiments in Lison and Kutuzov~\cite{lison2017redefining}.

%
%

    Next, we take a look at the results depending on task difficulty, which are presented in Table~\ref{tab:diff:over:td}.
    As expected, accuracy generally raises with decreasing task difficulty. But surprisingly, CBOW models perform very
    well on the hardest task category.
    On the hardest difficulty level, where intruders
    are semantically very similar to the other terms, the best results are at $76\%$. On the other hand \emph{w2v-ww12-i15-hs}, the best model on the analogy tasks, 
    only yields 52.7\% . On difficulty level 3 accuracy already goes up to 92\%. 
    All models solve the easiest category tasks with over 85\% accuracy, \emph{w2v-default} provides more than 95\%.


    For the Harry Potter dataset, the picture is similar and as expected. There are distinct differences and 5-20\% improvement
    from task difficulty class to task difficulty class among all models. The last line in Table~\ref{tab:diff:over:td}
    shows the difficulty level results for the HP dataset with a word2vec default model.

    \begin{table}[ht]
    \caption{ASOIF word intrusion dataset (uni-grams): Accuracy results regarding different levels of task difficulty.}
    \label{tab:diff:over:td}
    \begin{center}
    \begin{tabular}{l|l|l|l|l|l}
    \hline\noalign{\smallskip}
          Task Difficulty & 1 (hard)  & 2 (med-hard)  & 3 (medium) & 4 (easy) & AVG \\ 
            \noalign{\smallskip}\hline\noalign{\smallskip}
          Number of tasks:       & 2795 & 2795 & 2795 & 2795 & 11180 \\ 
            \noalign{\smallskip}\hline\noalign{\smallskip}
 w2v-default         & 74.60\%            & 69.70\%            & \textbf{92.24\%}  & \textbf{95.49\%}   & \textbf{83.01\%}  \\  
 w2v-ww12-i15-hs     & 52.74\%            & 70.77\%            & 79.75\%           & 88.30\%   & 72.89\%  \\  
 w2v-ww12-i15-ns     & 60.11\%            & 72.45\%            & 80.25\%           & 90.91\%   & 75.93\%  \\  
 w2v-CBOW            & 70.30\%            & 71.02\%            & 83.61\%           & 93.99\%   & 79.73\%  \\  

GloVe                & 66.40\%            & 56.21\%            & 89.27\%           & 93.13\%   &  76.25\%  \\ 
 fastText-default    & \textbf{75.78\%}   & 70.59\%            & 64.72\%           & 93.17\%   & 76.06\%  \\  
 fastText-ww12-i15-ns& 61.11\%            & \textbf{72.49\%}   & 81.68\%           & 91.59\%   & 76.72\%  \\  

 LexVec              & 66.22\%            & 70.05\%            & 62.07\%           & 85.36\%   &  70.93\%  \\ \hdashline 
 Baseline: PPMI      & 37.17\%            & 68.71\%            & 87.57\%           & 88.22\%   &  70.37\%  \\  

 HP: w2v-default     & 37.43\%            & 74.01\%            & 84.70\%           & 91.12\%   & 71.82\%  \\  

    \noalign{\smallskip}\hline
    \end{tabular}
    \end{center}
    \end{table}





The work of Sahlgren and Lenci~\cite{Sahlgren2016emnlp} includes an evaluation of term frequencies in distributional semantics on performance in
word similarity tasks. They find that low frequencies of task terms strongly reduce average accuracy, from around 60\% to 20\%-30\%
for word2vec models between medium and low frequencies. Medium frequency in their 1B token corpus is 800-16K occurrences, low frequency therefore $<800$. 
As our corpora are by a factor of 100 smaller, most terms are in low frequency range of~\cite{Sahlgren2016emnlp}.
To analyze impact of term frequency in our small domain corpus, we set up six frequency bins, with term frequencies
as provided in Table~\ref{tab:freq}. We evaluate the effect on performance in the word intrusion task of (i) the average frequency of the four terms involved,
(ii) the frequency of the correct (gold) term/outlier, (iii) the frequency of the outlier selected by the algorithm (which is not necessarily the correct one).
To our surprise, the frequency of the correct outlier (ii) has very little impact on accuracy, whereas the found term (iii) has the strongest impact.
The effect of average frequency is in between. Table~\ref{tab:freq} shows the impact of the frequency bin of the found term (iii) on word intrusion accuracy
for the model \emph{w2v-ww12-i15-ns}. Results are provided for both book corpora, and for uni-gram and n-gram datasets. The values in parentheses represent the number of
\emph{found terms} in the respective bin.


    \begin{table}[ht]
    \caption{The effect of \emph{found term} frequency on word intrusion accuracy -- on the basis of model \emph{w2v-ww12-i15-ns}.}
    \label{tab:freq}
    \begin{center}
    \begin{tabular}{l|l|l|l|l|l|l}
            \hline\noalign{\smallskip}
          Frequency Bin & 1 & 2 & 3 & 4 & 5 & 6  \\ \hline
          Term frequency: & 0--20 & 21--50 &  51--100 & 101--500 & 500--1000 & $>$1000 \\ 
            \noalign{\smallskip}\hline\noalign{\smallskip}

\multirow{2}{*}{ASIOF (uni-gram)}   & 57.69\%   &  92.09\%  &   53.30\%   &    69.52\%  &    98.56\%  &  92.27\%   \\  
                                    &  (52)     &   (569)   &    (1499)   &     (5670)  &     (1047)  &   (2343)   \\ \hdashline     

\multirow{2}{*}{HP (uni-gram)}      & 59.94\%   &  69.13\%  &   50.84\%   &    78.82\%  &    74.34\%  &  86.44\%   \\  
                                    &  (362)    &  (1166)   &    (1196)   &    (4199)   &     (686)   &   (811)   \\ \hdashline  

\multirow{2}{*}{ASIOF (n-gram)}     & 71.70\%  &  54.42\%  &   74.45\%    &    87.35\%  &   100.00\%  & 100.00\%   \\  
                                    &  (364)   &  (373)    &    (278)     &    (435)    &    (315)    &  (235)   \\  \hdashline 

\multirow{2}{*}{HP (n-gram)}        & 13.31\%  &  19.00\%  &   23.21\%    &    90.74\%  &   100.00\%  &  96.59\%   \\  
                                    &  (676)   &  (321)    &    (448)     &    (162)    &    (78)     &   (235)   \\  \hline   

\textbf{Average}                    & 50.66\%  &  58.66\%  &   50.45\%    &    81.61\%  &    93.23\%  &  93.83\%  \\  

    \noalign{\smallskip}\hline
    \end{tabular}
    \end{center}
    \end{table}

In a nutshell, if Gensim selects a term as outlier which has a frequency $>500$ (bins 5 and 6), then
the result is correct in over 95\% of cases. On the other hand, for frequencies $<100$ (bins 1, 2 and 3), the level of accuracy and 
trust in the result is very low.

Finally, we discuss the results on the n-gram datasets for the \emph{word intrusion} task.
For both datasets, \emph{fastText} is the clear winner.
The accuracy is at $89.5\%$ for the ASOIF data, this is a wide gap to the second
best performing model, which is \emph{w2v-ww12-i15-ns-ngram} at $80.8\%$.
The picture is similar for the HP dataset, although accuracy is lower in general.
\emph{fastText} uses subword information in vector construction, which helps in finding
words which have some lexical overlap. 
For \emph{fastText} it is particularly easy to find the outlier in an 
input question such as \texttt{narrow\_sea jade\_sea castle\_black salt\_sea}. All other models
only rely on word context. 
In the easiest difficulty class, \emph{fastText} solved the word intrusion tasks 
with about 98\% accuracy for the ASOIF dataset. 

Furthermore, the \emph{PPMI} baseline performed well in the n-gram setting, the accuracy
is on level with many of the word embedding models, except \emph{fastText}.

\subsubsection{Discussion of Results}

As we have seen from low accuracy on the analogies tasks, the problem is hard to solve 
with word embeddings alone. The difficulty stems on the one hand from the fact that relations between (for example) persons are mostly manifold and complex, and they change over time.
And on the other hand, the search space includes the complete model vocabulary. We understand our work on this task as providing a baseline for more elaborate future work.

For example, instead of the offset method, a classifier can be trained for relation detection, as in Gupta et al.~\cite{Gupta2017starsem}.
Another line of attack could be the reduction of the search space, eg. annotating the text with NLP tools, and filtering for matching word categories, etc.
Furthermore, grounding the terms in knowledge bases like DBpedia would allow the utilization of extensive structured information -- but would
also raise system complexity by far and limit the approach to literature extensively covered in Wikipedia. 
Another way to leverage structured data would be the attempt to combine regular embeddings with entity embeddings trained on Linked Data, see for example RDF2Vec~\cite{RistoskiP16}.

Regarding model performance on the analogy task, the tuned word2vec models (skip-gram) with an extended word window (eg: 12 words),
worked very well. In the analogy task a larger than default (5 words) word window improves performance a lot. As our experiments show, 
even very large windows (15--30 words), which cross sentence boundaries, can be useful.
On the other hand, for the word intrusion tasks the default word window provides best results in the evaluations, even in our setting of a very small corpus.
GloVe performs quite well on analogies, on word intrusion results were average. CBOW shows very good results on word intrusion. This is mostly in line with the experiments by Ghannay et al.~\cite{ghanny2016},esp.~with
regards to the performance of GloVe.
Despite its good performance on word intrusion, the experiments show that CBOW is not suitable for small corpora for the analogy task, which confirms Mikolov et al.~\cite{mikolov2013w2v}, but contradicts some
of the findings in Sahlgren and Lenci~\cite{Sahlgren2016emnlp}.

FastText is stronger in the word intrusion task than on analogies. This is especially true for n-grams,
where fastText clearly outperforms the competition with its ability to leverage subword information.
As state-of-the-art baseline method on the word intrusion task we applied PPMI. Most word embedding models beat PPMI or were at least on its level, 
except \emph{LexVec}. On the n-gram datasets, PPMI is on par with some embedding-based models for both book series datasets.

The difficulty levels on the word intrusion task fulfill their purpose to segregate the datasets into classes of decreasing challenge for the algorithms.
An interesting finding is that in our experiments CBOW-based models yield very good results on the hardest difficulty category. 
The result regarding the frequencies of the (found) term in the word intrusion tasks are in line with~\cite{Sahlgren2016emnlp}, if a term appears 
more than 100 times in the book text, the quality of embedding of the term improves drastically.

\section{Conclusions}
\label{sec:concl}

Both word embeddings and the application of computational linguistics in the digital humanities domain
gained a lot of attention recently. We combine the two in an evaluation of typical word embeddings tasks
like word intrusion or analogy on the well-known fantasy novel book series ``A Song of Ice and Fire'' (ASOIF) by George R.~R.~Martin, and ``Harry Potter'' (HP) by Joanne K.~Rowling.
First we train different word embedding models on the two corpora, and then create test datasets with the help of domain experts.
The datasets provide 16220 questions for the ASOIF book series, and 15254 for HP.
The analogies task is very hard given the small corpora and the complexity of relations, so we consider our work as a baseline for future evaluations. 
On the word intrusion task word embeddings typically outperform the state-of-the-art method of PPMI.
Furthermore, we study various related aspects, such as hyperparameter tuning with a grid search, the influence of
term frequencies of task terms on accuracy, and aspects specifically to named entities within distributional models.

The contributions of this work include: 
(i) The evaluation of the suitability of word embeddings trained on small corpora for the tasks at hand, 
including findings on which word embeddings tool works best in which scenario.
Some of the findings confirm previous research, for example that GloVe performs well on the analogy task,
 whereas word2vec is better suited for word similarity. A very interesting result is that for analogical reasoning 
very large word windows, potentially crossing sentence boundaries, are useful, whereas for word intrusion small word windows are beneficial. 
(ii) We manually created high-quality datasets in the digital humanities domain for two fantasy novels.
(iii) All created resources and the accompanying code base is shared for the purpose of replicating the experiments,
and esp.~for evaluating other approaches to the tasks at hand. The given task types are generally used in word embeddings
evaluation and therefore of broader interest and applicability. The resources include the datasets, the trained models, 
and the source code for dataset creation and evaluation. We tried to keep the system easy to comprehend and to extend
to new datasets or functionalities. 
(iv) Findings on the impact of term frequency of task terms on accuracy, and esp.~that the frequency of the term picked by the algorithm,
and not of the correct result matters most.

There are many possible lines of future work, here we will mention a couple of aspects:
(i) As corpus size has a big influence on statistical models, one obvious direction of work is the inclusion of external sources
like Wikis and fan sites, but this will also introduce many problems and biases. 
More interesting and fruitful might be using information generated by standard NLP tools on the book corpora, eg. to filter for word categories;
or to attempt linking the dataset terms to structured knowledge bases like DBpedia; or finally investigate the combination of current models with
entity embeddings such as RDF2Vec~\cite{RistoskiP16}. 
(ii) In the book corpora we face issues of ambiguity (eg. Jon Arryn vs Jon Snow) and multiple names of an entity (Dany and Daenerys),
which also have impact on the trained language models. In future work we will experiment with techniques for entity linking and disambiguation, 
and of co-reference resolution. 
(iii) Studying the details of relations between entities in domain-specific distributional models, and instance-of relations of entities to kinds.
(iv) Finally, relations, esp.~between characters, evolve over time, 
therefore a more fine-grain and temporal analysis may uncover these evolution aspects.

\section*{Acknowledgements}
This work was supported by the Government of the Russian Federation Grant 074-U01 through the ITMO Fellowship and Professorship Program.
Furthermore, the article was prepared within the framework of the Basic Research Program at the National Research University Higher School of Economics (HSE) and supported within the framework of a subsidy by the Russian Academic Excellence Project '5-100'. 
This work was also supported by RFBR grants 16-01-00583, and 16-29-12982.

\bibliographystyle{splncs03}      
\bibliography{our_bib}   

%
\end{document}